\documentclass[review]{elsarticle}

\usepackage{amsmath}
\usepackage{amsfonts}
\usepackage{xcolor}
\usepackage{booktabs}
\journal{Journal of \LaTeX\ Templates}

\begin{document}

\begin{frontmatter}

\title{Who's Afraid of Adversarial Transferability?}

%% Group authors per affiliation:
\author{Ziv Katzir\corref{mycorrespondingauthor}}
\cortext[mycorrespondingauthor]{Corresponding author}
\ead{zivka@post.bgu.ac.il}
\author{Yuval Elovici}
\address{Department of Software and Information Systems
Engineering, Ben-Gurion University of the Negev, Be’er Sheva 8410501,
Israel}

\begin{abstract}
Adversarial transferability, namely the ability of adversarial perturbations to simultaneously fool multiple learning models, has long been the "big bad wolf" of adversarial machine learning. Successful transferability-based attacks requiring no prior knowledge of the attacked model's parameters or training data have been demonstrated numerous times in the past, implying that machine learning models pose an inherent security threat to real-life systems. However, all of the research performed in this area regarded transferability as a probabilistic property and attempted to estimate the percentage of adversarial examples that are likely to mislead a target model given some predefined evaluation set. As a result, those studies ignored the fact that real-life adversaries are often highly sensitive to the cost of a failed attack. We argue that overlooking this sensitivity has led to an exaggerated perception of the transferability threat, when in fact real-life transferability-based attacks are quite unlikely. By combining theoretical reasoning with a series of empirical results, we show that it is practically impossible to predict whether a given adversarial example is transferable to a specific target model in a black-box setting, hence questioning the validity of adversarial transferability as a real-life attack tool for adversaries that are sensitive to the cost of a failed attack. 
\end{abstract}

\begin{keyword}
Adversarial transferability
\MSC[2010] 00-01\sep  99-00
\end{keyword}

\end{frontmatter}

%%\linenumbers

\section{Introduction}
It has long been known that adversarial perturbations that mislead a given learning model often mislead other models as well. This is true even when the other models are based on different algorithms and trained with independent datasets \citep{szegedy2013intriguing}. This property, which is commonly referred to as \textit{adversarial transferability} \citep{papernot2016transferability}, has been the source of much concern for  machine learning researchers and practitioners. On the face of it, transferability allows adversaries to successfully attack any machine learning-based system with no prior knowledge of the underlying models or their training procedures. As such, any real-life machine learning-based application is at risk of being manipulated by adversarial attacks.

The classic recipe for a transferability-based attack is surprisingly straightforward and has successfully been used multiple times in the past \citep{szegedy2013intriguing, papernot2016transferability, inkawhich2019feature, sah2020adversarial, wu2020boosting}. The adversary first trains a surrogate machine learning model using some publicly available data and state-of-the-art method; then uses known white-box attack algorithms to find adversarial examples, given the surrogate model; and finally, uses the resulting adversarial examples against the target model. Surprisingly enough, the results of prior studies show that around 30-70\% of the adversarial examples employed using this simple framework are able to mislead the target model. 

However, it is important to note that all prior transferability related research studied this phenomenon from a probabilistic perspective. Past works have all attempted to estimate the portion of adversarial examples created using one model that will remain effective against a target model. However, to the best of our knowledge, no attempt has been made to deterministically predict whether a specific adversarial example will be able to mislead a given target model. 

We argue that by adopting a probabilistic approach, previous works have overlooked a key aspect of real-life attacks - namely that adversaries are often highly sensitive to the consequences of a failed attack, and that this sensitivity greatly influences their choice of attack methods. 

For instance consider a state sponsored cyber espionage. Such operations often take many months, or even years, to plan and execute, and the attackers in this case are extremely sensitive to the consequences of being discovered. As such the attackers are willing to invest significant effort in ensuring that their true identity remains secret, even if the attack itself is discovered. Furthermore, similarly high effort is placed on protecting the attack infrastructure and methods so they can be used repeatedly. As a result, any attack method that cannot reliably protect the attacker's identity or assets, will likely be abandoned without being used in real life.
As a second example consider an attacker that aims to cross an airport without being detected by the airport's face recognition system. In order to do so the attacker might try to change his/her appearance, however since the attacker does not know which facial recognition system is used at an airport, he/she might want to leverage adversarial transferability. That is, the attacker might try to fool a surrogate facial recognition system, assuming that the adversarial attack will successfully fool the airport's system as well. In this case, the results of a failed attacker might be devastating from an attacker's point of view. Hence he/she would strive for certainty regarding the potential results of the operation, and as a by product for absolute certainty with respect to underlying adversarial transferability. Even an 80\% probability that the adversarial attack will successfully transfer to the airport's facial recognition system is unacceptable in this case. The attacker will not risk entering an airport knowing that there is a 20\% chance of detection. In this example the attacker needs absolute certainty regarding the success of the entire operation and thus absolute certainty regarding the attack's transferability from the surrogate model to the model used in the airport. 

In this work, we study transferability and examine its applicability given the real-life constraints mentioned here. In order to do so, we present a simple theoretical model to explain transferability itself and then explore the limits of transferability in real life through a series of empirical experiments. We show that transferability is solely dependent on the structure of the decision boundaries of both the surrogate and target models. Then we demonstrate that the nature of those boundaries makes it impossible to predict whether a given adversarial example will mislead the target model unless the adversary has clear knowledge of the target model's inner parameters. Such white-box knowledge of the attacked model's parameters is unrealistic in the vast majority of real life cases and therefore we conclude that when an attacker is sensitive to the cost of failure and requires high certainty on an attack's success, transferability becomes an invalid tool. As a result we believe that the true risk of transferability to real life systems is far lower than commonly believed.

\section{Understanding Adversarial Transferability}
In this section, we study the roots of the transferability phenomenon itself and by doing so, provide a solid basis for an analysis of its capabilities and limitations.

We start by noting that adversarial examples themselves are the result of the structure of the relevant classifier's decision boundaries, and we use this understanding in order to present a simple explanation for adversarial transferability. We claim that transferability is solely dependent on the position of the decision boundaries of the target and surrogate model with respect to the adversarial example and provide a simple mathematical proof to support that claim.

Finally, we discuss the underlying factors that define a classifier's decision boundaries. This discussion serves as the basis for our follow-up attempts to predict the transferability of a given adversarial point.

\subsection{Adversarial Examples Are Defined by a Classifier's Decision Boundaries}
Informally, adversarial examples are incorrectly classified data points that reside in proximity to some other, correctly classified, input. However, there is a pitfall in this informal definition. The terms 'incorrectly' and 'correctly' imply the existence of an all-knowing oracle that can provide the correct class label for any given input. Such an oracle clearly does not exist in most practical cases, and therefore past research has used a variety alternative definitions for adversarial examples.

In the context of this work we follow the most commonly used definition for adversarial examples in past studies: Given a classifier $f\left(\cdot\right)$, a correctly classified input $x$, a perturbation vector $\delta$, and a maximal perturbation distance $\varepsilon$, a data point ${x}' = x + \delta$ \;is considered adversarial when - 
\begin{equation}
\label{eq:1}
\begin{aligned}
f\left ( {x}' \right ) \neq f\left (x \right)\\
s.t.  \;\;\left\|\delta\right\| \leq \varepsilon
\end{aligned}
\end{equation}
In a recent study, this definition was called 'exact-in-the-ball' \citep{gourdeau2019hardness}, as it reflects the intuition that class label changes should not happen within a small radius around correctly classified inputs or, alternatively, that a minimal generalization sphere of radius $\varepsilon$ should surround any correctly classified input. This definition implicitly associates adversarial examples with the shape of the attacked model's decision boundaries. However, the existence of adversarial examples and the large number of different techniques available for finding them \citep{szegedy2013intriguing, goodfellow2014explaining, papernot2016limitations, kurakin2016adversarial, moosavi2016deepfool, carlini2017towards} clearly prove the shortcoming of our intuition with respect to the structure of those decision boundaries. In recent years it has become quite clear that class label changes can occur within a small distance from a correctly classified input \citep{zhang2016understanding, shafahi2018adversarial, shamir2019simple} and that the structure of the decision boundary in proximity to correctly classified input is far from being spherical. 

\subsection{Defining Adversarial Transferability}
Adversarial transferability in neural networks was first discovered in \citep{szegedy2013intriguing}. Formally, an adversarial example ${x}' = x + \delta$ is transferable from a surrogate model $f_{S}$ to a target model $f_{T}$ if and only if
\begin{equation}
\label{eq:2}
\begin{aligned}
f_{S}\left({x}' \right )\neq f_{S}\left(x \right )\rightarrow f_{T}\left({x}' \right )\neq f_{T}\left(x \right )
\end{aligned}
\end{equation}

Transferability can further be subdivided into targeted and non-targeted transferability. In the case of targeted transferability, the two classifiers assign the same class label to the adversarial example, whereas with non-targeted transferability, the target classifier is fooled by the adversarial example but may assign a different class label than the one assigned by the surrogate classifier. For the sake of notation convenience in later sections of the paper, we now define two indicator functions for the targeted and non-targeted transferability, respectively: 
\begin{equation}
\label{eq:3}
\begin{aligned}
T_{T}\left ( f_{T}, f_{S}, x, {x}' \right ) = \left\{\begin{matrix}
1 & f_{T}\left({x}' \right) \neq f_{T}\left( x \right )\;\cap\; f_{T}\left({x}' \right ) = f_{S}\left( {x}' \right ) \\ 
0 & otherwise
\end{matrix}\right.
\\
T_{N}\left ( f_{T}, f_{S}, x, {x}' \right ) = \left\{\begin{matrix}
1 & f_{T}\left({x}' \right) \neq f_{T}\left( x \right )\;\cap\; f_{S}\left(x \right ) \neq f_{S}\left( {x}' \right ) \\ 
0 & otherwise
\end{matrix}\right.
\end{aligned}
\end{equation}
When $f_{T}$ and $f_{S}$ are clear from the context of the text, we often use the shorthand notation, $T_{T}\left({x}' \right), T_{N}\left({x}' \right)$, omitting the explicit classifier names.

\subsection{Transferability Is Also Determined by Decision Boundaries}
In \citep{papernot2016transferability}, the authors suggested, for the first time, that transferability is the result of similarity in the decision boundaries of the two classification models. This rather intuitive explanation was empirically supported through an algorithm that was able to increase the observed transferability by increasing the boundary alignment between the two models.

Later studies have questioned the boundary similarity hypothesis, pointing out two interesting properties of transferability. In \citep{liu2016delving}, it was shown that non-targeted transferability rates are different than targeted ones. This, however, seems to contradict the boundary similarity assumption, since when the decision boundaries of the two models are similar to one another, one could expect the two models to produce identical classification decisions as well. Then \citep{wu2018understanding} showed that different transferability success rates are observed when switching the roles of the target and surrogate models. This asymmetry also seems to contradict the boundary similarity assumption, as similarity is inherently a symmetrical relation. Together, those two observations led researchers to look for alternative explanations for the transferability phenomenon. In \citep{wu2018understanding}, it was argued that boundary similarity cannot fully explain the nuances of transferability and that model specific properties, such as network architecture or gradient smoothness, provide a better explanation. A similar approach was taken in \citep{demontis2019adversarial}, where transferability was associated with the target model's susceptibility to attacks and the complexity of the surrogate model. 

We believe that the apparent contradiction presented here is in fact artificial. We claim that transferability is solely attributed to the decision boundaries of the two classifiers involved, which are in turn shaped by various model-specific properties. A simple, intuitive explanation for the first part of our claim is provided here, while the factors that influence the decision boundaries themselves are discussed later in the paper.

In order to show that decision boundaries are solely responsible for transferability, we focus on a single arbitrary adversarial example as opposed to some global sense of boundary similarity. Focusing on individual adversarial examples makes it easy to phrase our claim as follows: Let ${x}' = x + \delta$ denote an adversarial example against $f_{S}$. From \eqref{eq:1}, we get that $f_{S}\left(x\right) \neq f_{S}\left({x}'\right)$, meaning that $x$ and ${x}'$ reside on the opposite sides of some decision boundary, given $f_{S}$. Then, from \eqref{eq:2}, we get that ${x}'$ is transferable to $f_{T}$ if and only if $x$ and ${x}'$ reside on different sides of one of the decision boundaries in $f_{T}$ as well. That is, transferability is determined by the position of the decision boundaries of $f_{T}$ with respect to $x$ and ${x}'$, as we aimed to prove.

Note the subtle, yet crucial, difference between the simple proof provided here and the previous notions of 'boundary similarity': Our proof deals with a single input vector $x$ and a single adversarial example ${x}'$. Indeed, when the decision boundaries of the two models are highly aligned, there is a better chance of transferability. However, the transferability of each individual adversarial example is independent of all others. As a result the observed transferability is highly dependent on one's testing dataset, leading us to realize that different test point choices can potentially result in vastly different transferability rates for the same two classification models. This is particularly true given the complex high-dimensional nature of the decision boundaries of modern neural networks. 

\subsection{Decision Boundaries Are Determined by the Data, Algorithm, and Luck}
The decision boundaries of a given classifier are typically determined during its training process using some form of empirical risk minimization (ERM). In this process, the empirical error over some finite testing set is used as a proxy for the classifier's overall generalization error, and the training process attempts to minimize this test error to the extent possible. As a result, decision boundaries are jointly determined by the properties of the training algorithm and the distribution of the data used during training.

In some cases, such as in the case of artificial neural networks, a third factor influences the structure of decision boundaries as well. In those cases, training is performed by initializing model parameters randomly, and then using some iterative numeric optimization method in order to approximate the optimal decision boundary configuration. Therefore, the final set of model parameters and, as a by product, the final decision boundary configuration, is partially dependent on the initial random choice of parameter values. Putting it all together, we conclude that decision boundaries are dependent on the training data, learning algorithm, and some random choice of the initial parameters. 

The learning algorithm defines the overall general shape of the boundary surfaces; for example, linear SVMs define a single high-dimensional plane, decision trees and random forests split the input space into hypercubes that are associated with different class labels, nearest neighbor-based classifiers form sphere-like subspaces, and finally, deep neural networks form highly complex decision boundaries.

Training data shapes the classifier's decision boundaries in two complementary ways. First, it provides anchors in the input space for which the ground true class label is known, in many cases leading the training process to try to maximize the generalization margins around those training points. Secondly, the absence of training data from some parts of the input space also influences the resulting decision boundaries. When input dimensionality increases, the training data gradually becomes sparser, until it can no longer ensure equal generalization margins in all directions and for all of the different training points. Consequently, ~\citep{simon2019first} proved that the susceptibility of classification models to adversarial examples increases proportionally to the square root of their input dimensionality. Data sparsity allows decision boundaries to contain 'cavities,' which in turn make the classifier susceptible to adversarial examples.

Random choice is eventually responsible for shaping the decision boundary when the training data does not provide statistical support. Using deep neural networks as an example, when training data becomes sparser, the gradient signal resulting from changes in some regions of the input domain becomes weaker, and the training process is unlikely to update the initial choice of random parameter values. Hence, the boundary layout in those input subdomains will solely be determined by the initial random choice of the model parameters. In Section \ref{subsection:experiment1}, we empirically demonstrate how different random choices dramatically influence the structure of the decision boundary and thus the transferability of specific adversarial examples.

\section{Predicting Transferability in Black-Box Conditions is Practically Impossible}
From an attacker's point of view, transferability is only interesting if it can be controlled, or at least predicted, without prior knowledge of the target model's internal parameters. The attacker must know, in advance, whether a specific adversarial example will fool the target classifier or not. However, as transferability is determined by decision boundaries, predicting it is equivalent to predicting the target model's decision boundary position with respect to some adversarial input.

With that in mind, our next goal is to understand the extent to which an attacker can predict the target model's decision boundary position under black-box conditions. When approaching this task, we should pay attention to the amount of influence the attacker might have on the various factors that determine the boundary's position. The attacker might have some knowledge with regard to the algorithm and training data used to create the target model but will clearly have no influence over the initial random choice of model parameters. 

\subsection{Experiment I - Transferability Prediction for a Knowledgeable Adversary}
\label{subsection:experiment1}
We start by attempting to predict transferability based on the strongest attacker model that can still be considered 'black-box' - we allow the attacker full knowledge of the target model's training algorithm, architecture, and training set, and keep only only the attacked model parameter values secret. In this setting, the only difference between the target and surrogate models is that they use a different set of initial random parameter values. 

In order to assess the attacker's capability in this setting, we use the following basic experimental setting.
\begin{enumerate}
    \item First, we trained a convolution neural network on the MNIST \citep{lecun1998gradient} dataset, which will be used as our surrogate model. We denote this classifier as $f_{S}$
    \item Then, we randomly picked 1,000 data points (i.e., images) from the MNIST testing set and identified 100 independent adversarial examples based on each of those points, for a total of 100,000 adversarial examples altogether. 
    Let $X^{'}$ denote the complete set of adversarial examples, and let $x_{p,d}^{'} \in X^{'}$ denote a specific example from within that set, where $p$ is the source point index  (out of 1000), and $d$ is the index of the adversarial example (out of 100).
    
    We use the HopSkipJumpAttack method \citep{chen2019hopskipjumpattack} for this purpose, as two of its core features make it particularly suitable for our needs: a. It allows multiple different adversarial perturbations to be identified for a given source point, and b. It operates by "traversing" the attacked classifier's decision boundary, so that the resulting perturbations reside very close to the model's decision boundaries.
    
    Notably, HopSkipJump was originally suggested as a black-box attack method, however in the context of this work, we intended to use it in a white-box setting against our surrogate model. We therefore implemented our own white-box variant of this method, giving up the black-box nature, but gaining improved computation speed (both the HopSkipAttack method and our own implementation details are discussed in \ref{appendix:A}).
    \item Third, we trained $N_{s} = 100$ additional classifiers denoted $f_{T_{j}} \;\; j \in \{0..N_{s} \}$ to be used as our attack targets. The target models were trained using the same algorithm, network structure, and dataset used previously for training the surrogate model. On their initial random model weights is different.
    \item Finally, we tested the transferability of each adversarial data point to each of the 100 target models and computed the transferability expectation for each adversarial example - 
    \begin{equation}
    \label{eq:4}
    \begin{aligned}
    \mathbb{E} \left[ T \left(f_{T_{j}}, f_{S}, x_{p}, x^{'}_{p,d} \right ) \right] = \frac{\sum_{j=0}^{N_{s}} T\left(f_{T_{j}}, f_{S}, x_{p}, x^{'}_{p,d} \right )}{N_{s}}
    \end{aligned}
    \end{equation}
    
    In addition we also measured the mean standard deviation of the transferability expectation per source point $\frac{\sum_{p=0}^{1000} \sigma_{p}}{1000}, \;\;\sigma_{p} = \sigma\left(\mathbb{E}\left[T\left( x^{'}_{p,d}\right ) \right ] 0\leq d\leq 100\right )$ as well as the standard deviation of the transferability expectation measured across all adversarial examples in $X^{'}$
\end{enumerate}

\begin{figure}
	\centering
		\includegraphics[width=\textwidth]{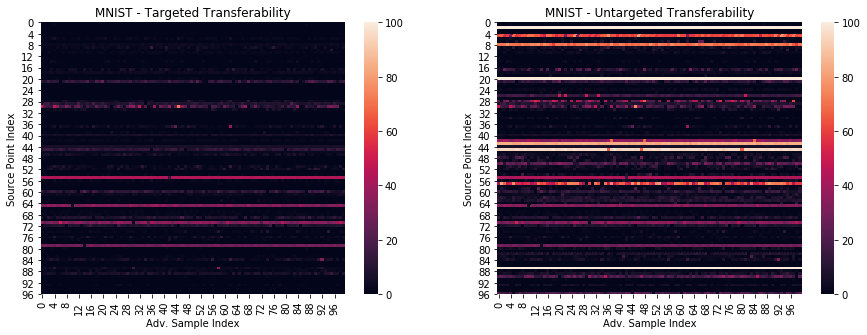}
	\caption{Targeted and non-targeted transferability expectation for the MNIST dataset}
	\label{fig:MNIST Transferability}
\end{figure}

The results of our first experiment are illustrated in Figure \ref{fig:MNIST Transferability}. The left hand heat-map diagram illustrates targeted transferability expectation, while the right hand one illustrates non-targeted transferability expectation. In both heat-maps the vertical axis enumerates the different source points, the horizontal axis enumerates the different adversarial perturbations created for each source point, and the color intensity represents the transferability expectation $\mathbb{E} \left[ T \left ( x^{'}_{p,d} \right) \right ]$. For brevity, the diagrams in Figure \ref{fig:MNIST Transferability} illustrate only 100 of the 1,000 source points used in this experiment; similar results are obtained when visualizing other source points from the 1,000 randomly selected data points. 

Several key observations become immediately clear when analyzing the results of our first experiment:
\begin{itemize}
    \item \textbf{Transferability is a local phenomenon -} By examining Figure \ref{fig:MNIST Transferability}, one can easily see that the color variance observed between different adversarial examples from the same source point (i.e., within a given heat-map row) is far smaller than the variance between adversarial examples associated with different source points. Similarly, the mean standard deviation of the non-targeted transferability expectation per source point $\frac{\sum_{p=0}^{1000} \sigma_{p}}{1000}$ is 0.02, while the standard deviation computed over all perturbations of all source points combined is 0.18.  This indicates that transferability (be it targeted or not) is local, in the sense that it is highly dependent on the source point from which the adversarial examples are derived.
    
    Looking at individual source points (horizontal rows in the heat-map), we can see that in many cases, high variance can also be seen within a single row. This further reinforces the notion of the 'locality' of adversarial examples, implying that transferability should be measured for each adversarial example independently of all others.
    \item \textbf{Transferability rates are far from homogeneous -} A second observation is that vastly different mean transferability expectation rates are associated with different source points. A high transferability expectation rate is observed for very few source points, while in the majority of cases, transferability expectation is low or even negligible.
    \item \textbf{Targeted transferability is a rare phenomenon -} Generally speaking, we can see that the color intensities in the left hand diagram of Figure \ref{fig:MNIST Transferability} are considerably lower than in the right hand diagram. The mean non-targeted transferability expectation rate in our experiment measures 0.08, while the mean targeted transferability expectation stands at 0.02. While targeted transferability is a more specific phenomenon, and one could expect a lower targeted transferability rate compared to the non-targeted rate, visualizing both heat-map diagrams side by side makes it clear that strong targeted transferability is hard to find.
    \item \textbf{Perfect transferability is even rarer -} None of the source points reached targeted transferability expectation above 0.6, and even for non-targeted transferability, very few source points are associated with an average transferability expectation of 0.8 or higher. This heightens the potential risk from an attacker's perspective.
\end{itemize}

The abovementioned observations seem to support our main claim in this work, i.e., they highlight the local nature of adversarial transferability and suggest that predicting the transferability of individual adversarial examples is a difficult task.

\subsection{Experiment II - Examining the 'Locality' of Adversarial Transferability}
\label{subsection:experiment2}
Encouraged by the results of the first experiment, we conducted a second one using the CIFAR-10 dataset \citep{krizhevsky2009learning}. Our goal in this experiment was twofold. First, we wanted to confirm our previous results with a richer dataset and a more complex classifier network. Second, we wanted to gain a more fine-grained understanding of the local nature of adversarial transferability. We already saw that multiple perturbations against a single source point demonstrate comparable transferability expectation rates, however we wanted to determine whether they also behave similarly in the context of specific target models. That is, will different adversarial perturbations against a single source fool the same set of target classifiers?

Our second experiment largely followed the setup of the experiment outlined set in Section \ref{subsection:experiment1}. 
\begin{enumerate}
    \item We trained 101 VGG 19 \citep{simonyan2014very} classification neural networks on the CIFAR-10 training dataset. One of the trained models was designated as the surrogate model (denoted $f_{S}$), and the remaining 100 models served as targets for our adversarial attacks (denoted $f_{T_{j}}, j\in\left\{0..99\right\}$).
    \item We then randomly selected 100 source points from the CIFAR-10 test set and created 100 independent adversarial examples for each of those source points. The result was a set of 10,000 adversarial examples. Following the notation set previously we use $X^{'}$ to denote the full set of adversarial examples and $x^{'}_{p,d} \in X^{'}$ to denote a specific adversarial example where $p$ is the index of the source point and $d$ is the index of the adversarial perturbation against that source.
    As before, we used the HopSkipJump attack method to identify relevant adversarial examples.
    \item Finally, we tested the transferability of each of the 10,000 adversarial examples created in the previous step and measured the transferability expectation. We also measured the mean standard deviation of the transferability expectation per source point $\frac{\sum_{p=0}^{100} \sigma_{p}}{100}$ as well as the standard deviation of the transferability expectation measured across all adversarial examples in $X^{'}$.
\end{enumerate}

\begin{figure}
	\centering
		\includegraphics[width=\textwidth]{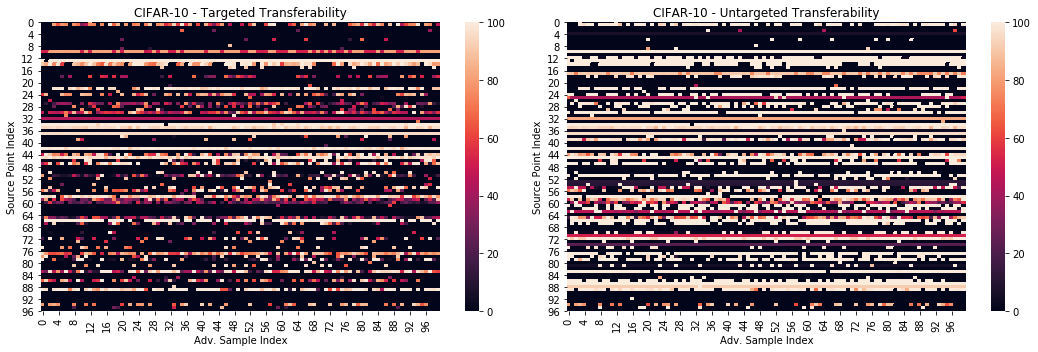}
	\caption{Targeted and non-targeted transferability expectation for the CIFAR-10 dataset}
	\label{fig:CIFAR transferability}
\end{figure}

\begin{figure}
	\centering
		\includegraphics[width=\textwidth]{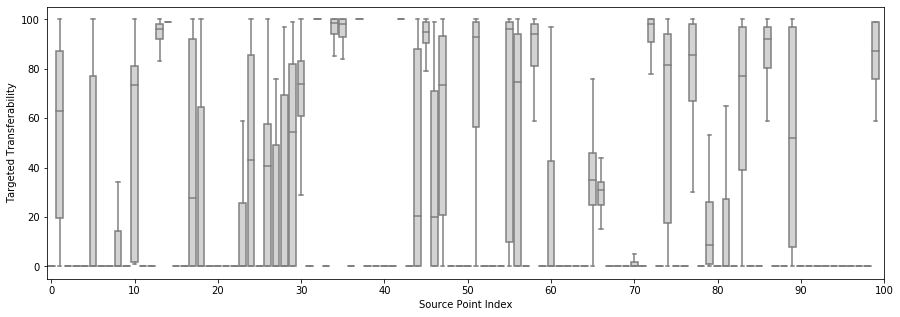}
	\caption{Transferability expectation distribution for individual source points in the CIFAR-10 dataset}
	\label{fig:CIFAR transferability distribution}
\end{figure}

The results of this experiment are illustrated in Figures \ref{fig:CIFAR transferability} and \ref{fig:CIFAR transferability distribution}. Figure \ref{fig:CIFAR transferability} illustrates the targeted and non-targeted transferability expectation rates observed in our experiment; as can be seen, the results are similar to those of the first experiment using the MNIST dataset. The left hand heat-map diagram illustrates targeted transferability expectation, while the right hand one presents the non-targeted results. In each heat-map, the vertical axis enumerates the different source points, the horizontal axis enumerates the different adversarial perturbations created for each source point, and the color intensity represents the transferability expectation.

Analyzing Figure \ref{fig:CIFAR transferability} provides us with the affirmation we sought. From the diagram it is clear that the results on the CIAR-10 dataset are similar to those observed for the MNIST dataset:
\begin{itemize}
    \item Each of the heat-map rows seems to have a typical mean color, indicating that transferability has a strong local tendency. This observation is again supported by comparing the mean standard deviation computed over the rows of the heat-map, with the standard deviation computed against all perturbations combined. The mean standard deviation computed over perturbations of the same source $\frac{\sum_{p=0}^{100} \sigma_{p}}{100}$ stands at  0.19, while the standard deviation computed over all 10,000 perturbations combined reaches 0.45.
    \item The targeted transferability expectation is again lower than the non-targeted transferability expectation. The mean non-targeted transferability expectation measures 0.36, while the targeted transferability expectation stands at 0.23, and
    \item Perfect transferability is again extremely rare.
\end{itemize}


After reaffirming the results of our previous experiment on the CIFAR-10 dataset, we moved on to try and improve our understanding of the interaction between source points and target classification models. Figure \ref{fig:CIFAR transferability distribution} presents a box plot of the targeted transferability expectation. Each bar in the box plot graphically illustrates the statistical distribution of the transferability expectation results associated with a single attack source point (that is, a single row in Figure \ref{fig:CIFAR transferability}). 
The horizontal line crossing each of the bars denotes the median value, the bottom and top edges of each colored bar denote the first and third quartiles respectively, and the lines extending outside of the colored box denote the minimal and maximal observed values. 
Tight boxes indicate that different perturbations derived from the same source point demonstrate similar transferability expectation levels, while wider boxes indicate that transferability expectation levels differ from one perturbation to another. Examining Figure \ref{fig:CIFAR transferability distribution}, we can see multiple cases where the median value is relatively high, but the first and third quartile marks are quite separated from one another. This indicates that even for a single source point, individual perturbations manage to fool different sets of target classifiers. That is, the boundary position with respect to the source point differs based on the direction of the perturbation vector. From an adversary's point of view, those results imply that predicting transferability for a given perturbation and target model is even more difficult than suggested by the results presented in previous subsections.

\subsection{Experiment III - The Role of the Surrogate Model}
\label{subsection:experiment3}
The first two experiments described above indicate that transferability is hard to predict in the context of a specific perturbation and a given target model. Using decision boundary terminology, we can say that the exact boundary position behaves in a somewhat chaotic way, with minor changes in the input factors leading to unpredictable transferability results.  Still, further exploration was needed in order to understand the effect of the surrogate model itself over the observed adversarial transferability. More specifically we wanted to assess the influence of the random choice of weights used for training the surrogate model over the adversarial transferability. That is, will there be differences in  the transferability observed for two surrogate models that were trained in a similar fashion?

Therefore, we extended the experiment scheme described in Section \ref{subsection:experiment2} slightly: 
\begin{enumerate}
    \item We trained 102 VGG 19 \citep{simonyan2014very} classification neural networks on the CIFAR-10 training dataset; two trained models were designated as surrogates (denoted $f_{S_{0}}, f_{S_{1}}$), and the remaining 100 served as target models (denoted $f_{T_{j}}, \;\;j\in\{0..99\}$).
    All 102 classifiers share the same network architecture and training dataset, differing only by the initial set of weights randomly chosen at the beginning of the training process.
    \item Once again, we randomly selected 100 source points from the CIFAR-10 testing set and used the HopSkipJump algorithm to identify adversarial perturbations. However, this time we created two sets of adversarial perturbations, one for each of the surrogate models. Each set included 100 independent adversarial perturbations for each source point (for a total of 10,000 adversarial examples in each set). 
    
    Following the notation from previous sections, let $X^{'}_{i}$ denote the set of adversarial examples associated with $f_{S_{i}}$ and let $ x^{'}_{i,p,d} \in X^{'}_{i}$ denote a specific adversarial example in that set, where $p$ is the source point index, and $d$ is the index of the adversarial example. Our goal was to understand whether the initial set of weights randomly chosen when training the surrogate model has any effect on the transferability observed. In order to do, so we used identical attack parameters for each matching pair of adversarial examples in the two sets ($X^{'}_{0}$ and $X^{'}_{1}$). As a result, any difference in the transferability of $x^{'}_{0,p,d}$ and $x^{'}_{1,p,d}$ is solely attributed to the random choice of weights used to train the two surrogate models.
    \item We tested the targeted and non-targeted transferability expectation from each set of adversarial examples to our 100 target models.
    For each adversarial example and destination model we recorded whether the classifier was misled by the adversarial example, and if so, whether the class label assigned to the adversarial example matched the desired target class. 
    \item Finally, we computed the portion of adversarial examples where $T \left( x^{'}_{0,p,d} \right) = T \left( x^{'}_{1,p,d} \right)$, i.e., the percentage of adversarial examples where the transferability's success is independent of the choice of the surrogate model. We refer to this measurement as the 'surrogate agreement' percentage.
    Given that the two surrogates share the same algorithm, architecture, and training data, and that identical attack parameters were used to create the two sets of adversarial examples, this measurement represents the effect of the random choice of weights in the surrogate model on the transferability expectation rate.
\end{enumerate}

\begin{figure}
	\centering
		\includegraphics[width=\textwidth]{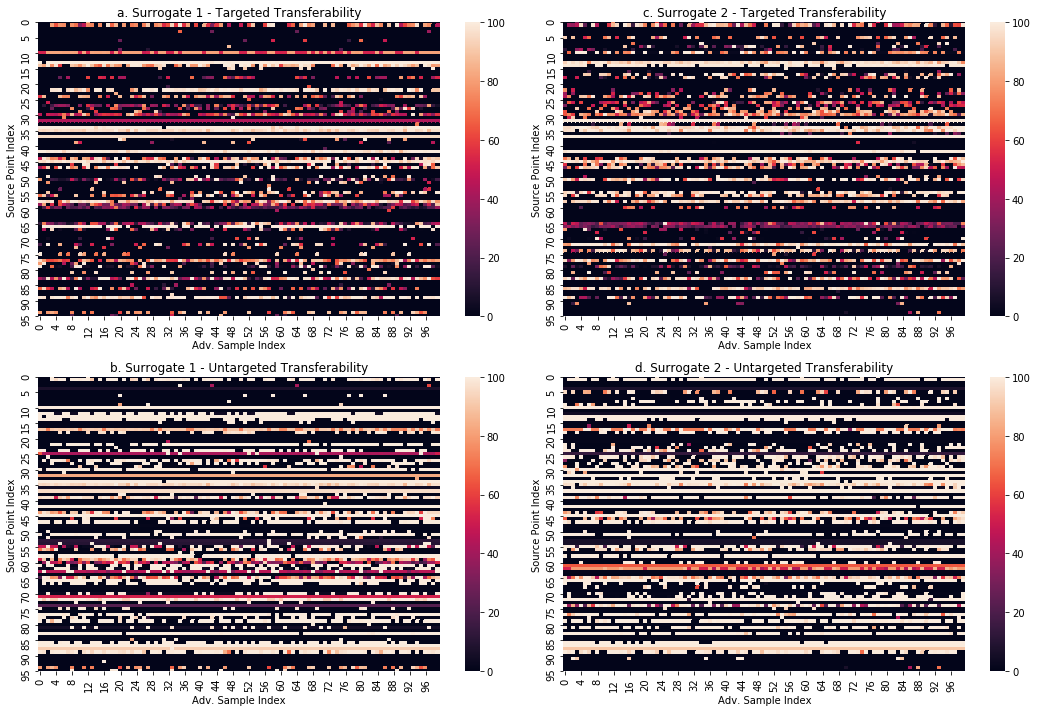}
	\caption{Transferability expectation for the CIFAR-10 dataset using two distinct surrogate models}
	\label{fig:CIFAR Transferability 2 surrogates}
\end{figure}

The results of this experiment are presented in Figure \ref{fig:CIFAR Transferability 2 surrogates} and Table \ref{table:surrogate-agreement}. Figure \ref{fig:CIFAR Transferability 2 surrogates} compares the transferability expectation estimated using the two different surrogate models. It includes four heat map diagrams, which are similar to the ones used in our previous experiments. The diagrams are organized in two rows and two columns. Heat maps labeled 'a', 'b' illustrate the transferability expectation for one surrogate model, while 'c' and 'd' illustrate the transferability expectation of a second surrogate. The top row of heat maps (i.e. heat maps 'a', 'c') illustrate the targeted transferability expectation for both surrogates, and the bottom row (i.e. heat maps 'b', 'd') illustrates the non-targeted transferability expectation. 

Careful examination of Figure \ref{fig:CIFAR Transferability 2 surrogates} reveals noticeable differences in the transferability expectation from one surrogate model to another. That is, examining specific pairs of matching perturbations for the two surrogates ($x^{'}_{0,p,d}$, $x^{'}_{1,p,d}$) we often observe that $\mathbb{E}\left[T\left(x^{'}_{0,p,d}\right) \right] \neq \mathbb{E}\left[T\left(x^{'}_{1,p,d}\right) \right]$. This observation is strengthened by the surrogate agreement levels reported in Table \ref{table:surrogate-agreement}. 

Although far from perfect, the overall surrogate agreement reported in Table \ref{table:surrogate-agreement} seems reasonably high, measuring 82\% for the targeted case and 80\% for the non-targeted case. While this is so, those results also mean that about 20\% of the transferability results are affected by the original choice of random weights used to train the surrogate model. Furthermore, one should also note the high number of samples for which zero transferability is observed for both surrogates. Those samples significantly bias the agreement results towards high-agreement rates. In order to eliminate that biasing effect, we exclude the adversarial examples that did not fool any of the target classifiers and report the surrogate agreement rate for the samples that managed to fool at least one target classifier for at least one surrogate. When doing so, the agreement rates drop to 14\% for the targeted case and 27\% for the non-targeted one. That is, the observed transferability is highly dependent on the surrogate model itself. Given that our surrogate models differ only based on their original choice of random weights, it is clear that an adversary cannot possibly attempt to predict the transferability of a given adversarial example using some in-lab surrogate setup.

All in all, the results of our third experiment provide additional empirical evidence to support our main claim - predicting the transferability of a specific adversarial example is practically impossible even when the adversary is aware of the target classifier's classification algorithm, training set, and model architecture.

\begin{table}[]
\centering
\begin{tabular}{lcc}
\hline
                             & \multicolumn{1}{l}{\begin{tabular}[c]{@{}l@{}}Targeted\\ Transferability\end{tabular}} & \multicolumn{1}{l}{\begin{tabular}[c]{@{}l@{}}Non-targeted\\ Transferability\end{tabular}} \\ \hline
Overall surrogate agreement  & 0.82                                                                                   & 0.8                                                                                      \\
Non-zero surrogate agreement & 0.14                                                                                   & 0.27                                                                                    
\end{tabular}
\caption{Targeted and non-targeted surrogate agreement}
\label{table:surrogate-agreement}
\end{table}

\subsection{Experiment IV - Transferability Prediction With No Prior Knowledge}
\label{subsection:experiment4}
After experimenting with a simplified attack model, we attempted a more realistic one in our fourth experiment with a more constrained attacker model. In this experiment, we assumed that the attacker is unaware of the algorithm used for training the target classifier but still allowed the attacker to have knowledge about the training dataset.
We repeated the general experiment scheme outlined in Section  \ref{subsection:experiment1}, training a CNN surrogate classifier based on the MNIST dataset, creating adversarial examples as before, and measuring the transferability expectation using a set of 100 target classifiers. This time, however, we used random forest (RF) based classifiers as our target models instead of CNNs. Using a CNN for the surrogate model and random forest for the target models, simulates the attacker's knowledge gap with respect to the internal details of the attacked classifier. In addition, the inherent 'randomness' of the random forest algorithm made it suitable for our experimental needs, as it creates slight variations between the decision boundaries of different target models.

\begin{figure}
	\centering
		\includegraphics[width=\textwidth]{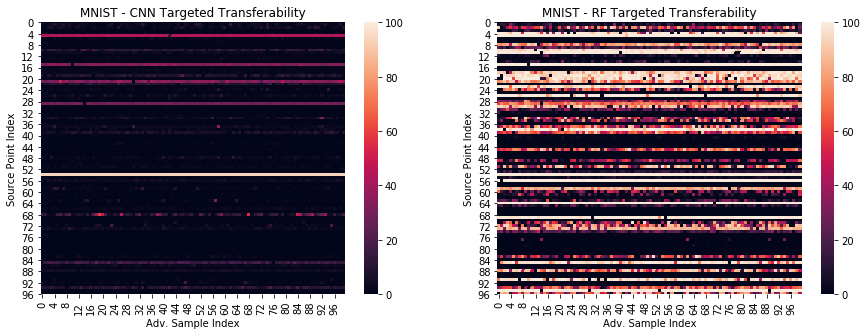}
	\caption{Transferability from a single surrogate model to 100 target CNN models compared to 100 random forest classifiers}
	\label{fig:MNIST CNN to RF Transferability}
\end{figure}

\begin{figure}
	\centering
		\includegraphics[scale=0.5]{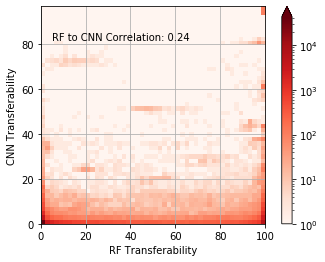}
	\caption{Comparing the transferability of the CNN and random forest target classifiers}
	\label{fig:MNIST CNN RF correlation}
\end{figure}

In this experiment, we compared the transferability expectation observed for the CNN-based classifiers as part of our first experiment to that observed with RF classifiers. The results of this experiment are illustrated in Figures \ref{fig:MNIST CNN to RF Transferability} and \ref{fig:MNIST CNN RF correlation}. Figure \ref{fig:MNIST CNN to RF Transferability} visualizes the targeted transferability expectation of the adversarial examples associated with 100 randomly chosen source points. The left hand heat-map diagram represents the transferability expectation computed over the 100 CNN classifiers from the first experiment, while the right hand diagram visualizes the transferability expectation computed over the 100 RF classifiers. In both heat-maps, the vertical axis enumerates the source point index, the horizontal axis represents the individual adversarial examples associated with each source point, and the color intensity represents the transferability expectation of each point.

Three main observations can be made based on an analysis of Figure \ref{fig:MNIST CNN to RF Transferability}:
\begin{itemize}
\item Perhaps the most immediate observation is that the color intensities in the right hand heat-map are significantly higher than those on the left. Correspondingly, the mean transferability expectation computed for the RF classifiers stands at 0.74 compared to just 0.08 for the CNN classifiers.
\item While the increased transferability associated with the RF classifiers might seem encouraging from an attacker's point of view, a deeper look at Figure \ref{fig:MNIST CNN to RF Transferability} reveals that the variance in transferability expectation in the case of RF-based classifiers is higher as well. The mean standard deviation computed per source point (i.e. the rows of the heat maps) was 0.13 for the RF target classifiers compared to 0.02 for the CNNs. Similarly, the overall standard deviation computed over all perturbations combined was 0.37 for the RF target classifiers and 0.18 in the case of the CNNs. This indicates that transferability in the RF case is even harder to predict than it was for the CNN targets.
\item The two heat-maps together seem to indicate a low correlation between the transferability expectation computed for the different types of target classifiers. Indeed, the Pearson correlation between the transferability computed using the different classifier types is 0.24. In practice this indicates that an attacker will be unable to predict the true transferability to a target model based on some in-lab experiment.
\end{itemize}
In order to obtain a better understanding of the correlation in transferability expectation, we used the heat map presented in Figure \ref{fig:MNIST CNN RF correlation}. Here the vertical axis denotes the transferability expectation computed for a given perturbation across the CNN classifiers, the horizontal axis denotes the transferability expectation computed for the same perturbation when using the RF classifiers as targets, and the color intensity indicates the number of adversarial examples that map to each coordinate in the diagram. Note that colors are presented using a logarithmic scale. This is done in order to compensate for the large differences in the values populating the heat map. 

Examining Figure \ref{fig:MNIST CNN RF correlation}, the low correlation between the transferability expectation observed for each of the classifier types is highly visible. The transferability computed across the CNN classifiers is practically independent of that computed across the RF classifiers. This implies that transferability to a given set of target classifiers cannot be used to predict the transferability to some other unknown model.

The results of this final experiment go hand in hand with those reported in previous sections. Transferability appears to be a fragile phenomenon that is highly dependent on the local structure of the decision boundaries of the two classifiers involved. Based on these results, we believe that it is practically impossible to predict whether a given adversarial example can be successfully transferred to some unknown target classifier. 
For a potential attacker, transferability's chaotic, unpredictable nature translates to low confidence and increased risk. As a result, we believe that the threat posed to real-life systems by adversarial transferability is much lower than previously thought, and in fact can probably be disregarded in cases where the attacker is sensitive to attack failure. 


\section{Conclusions}
In this work, we aimed to gain better understanding of the threat posed by adversarial transferability to real-life systems. More specifically, we sought to understand whether transferability can allow an attacker to successfully attack a target system, with no prior knowledge of its internals. 

Past research indicated that all machine learning based systems are inherently prone to transferability-based attacks, however all of the studies regarded transferability as a global, probabilistic phenomenon that affects all regions of a classifier's input domain equally. We believe that by taking this probabilistic approach, past research has overlooked the fact that attackers are often highly sensitive to the potential implications of a failed attack as well as the vast influence those concerns have on an attacker's choice to use one attack method or another in real life. Our goal, therefore, was to assess whether it is possible to predict the transferability of a given individual adversarial example to a specific target model without knowledge of its internal details. 

We started by examining the roots of adversarial transferability and showed that transferability is dependent solely on the structure of the decision boundaries of the two classifiers involved. Then, we explored the different factors that determine the structure of those decision boundaries and showed that the boundaries are defined by the training set, learning algorithm, and potentially the original choice of random model parameters at the beginning of the training process.

We then moved on to testing the predictability of transferability using various datasets and different levels of attacker knowledge. Our results show that transferability is highly sensitive to the structure of the decision boundaries in proximity to the perturbed input, and that those boundaries are unpredictable given minor changes to the training process or dataset of the relevant classifiers. As such it is practically impossible to predict whether a given adversarial example can fool a specific target model unless the attacker has full knowledge of that model's internal parameters. Given attackers' sensitivity regarding the potential implications of attack failure, we therefore conclude contrary to common belief, that transferability is not much of a threat to real-world systems. 
\bibliography{mybibfile}

\appendix
\section{Traversing Decision Boundaries in Search of Adversarial Examples}
\label{appendix:A}
Adversarial attack methods can largely be divided into two main categories: methods that traverse the classifier's loss surface and methods that explore the classifier's decision boundary itself \citep{fidel2020adversarial}. While in both cases the end result is very similar, the optimization approach used in each case is quite different. 
Given a valid source input to perturb, loss surface-based approaches typically use gradient-based optimization (such as stochastic gradient descent) to identify a perturbation that will maximize classification loss. Boundary traversal approaches, on the other hand, traverse the decision boundary itself, looking for the closest point to the source input.

The HopSkipJumpAttack method \citep{chen2019hopskipjumpattack} is a black-box, boundary-based attack that uses the attacked classifier as an oracle. This method uses a pair of input data points, a source point $x$ for which the adversary is looking to find an adversarial example, and a second input $\Tilde{x}_{0}$ belonging to the chosen target class. The algorithm operates by iteratively repeating the following steps until a relevant pertubation is found or the maximal number of iterations has been performed:
\begin{enumerate}
    \item In each iteration $t$, use the binary search algorithm to find the point at which the classifier's boundary surface and the hyperline connecting $x$ and $\tilde{x}_{t+1}$ intersect. Denote that intersection point as $x_{t}$.
    \item Use the Monte Carlo search algorithm, along with queries to the oracle, to 'map' the classifier's inferred class labels in proximity to $x_{t}$
    \item Based on the inferred class labels from the previous step, compute a linear approximation of the boundary at the intersection point.
    \item Compute $x_{t+1}$, the target point for the next iteration, by taking a small step perpendicular to the boundary's linear approximation
\end{enumerate}

The black-box nature of this algorithm, along with the Monte Carlo based gradient approximation, allows it to be used with non-differential classifiers as well. However, the black-box query based approach also results in increased computation times compared to other white-box approaches. In this paper, our interest was in the boundary traversal nature of the attack, with the intent of using it against our own, neural network-based classifier in a white-box fashion. Therefore, we  replaced steps 3 and 4 above with a simple white-box calculation of the boundary gradient at the intersection point. We also applied several code-based optimizations, which resulted in a considerable improvement in performance.

\end{document}